\documentclass{article}

\usepackage[dblblindworkshop, final]{neurips_2025}
\workshoptitle{AI That Keeps Up: Workshop on Continual and Compatible Foundation Model Updates (CCFM)}

\usepackage[utf8]{inputenc} 
\usepackage[T1]{fontenc}    
\usepackage{hyperref}       
\usepackage{url}            
\usepackage{booktabs}       
\usepackage{amsfonts}       
\usepackage{nicefrac}       
\usepackage{microtype}      
\usepackage{xcolor}         
\usepackage{multirow}
\usepackage{algorithm}
\usepackage{algpseudocode}
\usepackage{amsmath, amssymb}
\usepackage{graphicx}
\usepackage{placeins}   

\title{Per-Axis Weight Deltas for Frequent Model Updates}

\author{
  Stefan Kuyumdzhiev\\
  High School "Vasil Drumev"\\
  Veliko Tarnovo\\
  \texttt{stefan.kuiumdjiev@gmail.com} 
  \And
  Radostin Cholakov \\
  Stanford University\\
  \texttt{radicho@stanford.edu} 
}

\begin{document}

\maketitle

\begin{abstract}
Serving many task-specialized LLM variants is often limited by the large size of fine-tuned checkpoints and the resulting cold-start latency. Since fine-tuned weights differ from their base model by relatively small structured residuals, a natural approach is to represent them as compressed deltas. We propose a simple 1-bit \emph{delta} scheme that stores only the sign of the weight difference together with lightweight per-axis (row/column) FP16 scaling factors, learned from a small calibration set. This design preserves the compactness of 1-bit deltas while more accurately capturing variation across weight dimensions, leading to improved reconstruction quality over scalar alternatives. From a systems perspective, a streamlined loader that transfers packed deltas in a single operation per module reduces cold-start latency and storage overhead, with artifacts several times smaller than a full FP16 checkpoint. The method is drop-in, requires minimal calibration data, and maintains inference efficiency by avoiding dense reconstruction. Our experimental setup and source code are available at \url{https://github.com/kuiumdjiev/Per-Axis-Weight-Deltas-for-Frequent-Model-Updates}.
\end{abstract}

\section{Introduction}
Large foundation models continue to grow in size and computational demand, making both training and deployment increasingly resource-intensive \citep{kaplan2020scaling}. Once pre-trained, these models are often adapted to downstream tasks through fine-tuning. Depending on the setting, fine-tuning may involve updating all parameters with a supervised objective (full fine-tuning), applying low-rank updates as in LoRA \citep{hu2021lora} or other parameter-efficient fine-tuning methods ~\citep{houlsby2019adapters, ben-zaken2022bitfit, mahabadi2021compacter, dettmers2023qlora, zhang2023adalora, liu2024dora, kopiczko2024vera}, or reinforcement learning post-training, which can target either entire weight matrices or restricted subsets of parameters \citep{han2024peftsurvey}.
In cases where fine-tunes are represented as full weight updates, serving multiple variants remains a deployment challenge. Each fine-tuned checkpoint must be stored and loaded in its entirety, and switching between them requires keeping large weight tensors resident in GPU memory. This is particularly costly for inference providers that serve many users or domains simultaneously, and for continual adaptation settings where new model variants are introduced frequently \citep{zhang2024slora,sun2024punica}.
Yet weights of fine-tuned models are rarely far from their base counterparts. Across a variety of adaptation procedures, the resulting weight matrices tend to differ from the pre-trained model only by relatively small residuals, both in magnitude and in spectral structure \citep{liu2024bitdelta}. This suggests that storing a full checkpoint per fine-tune is wasteful: the information required to recover the specialized model lies in a compact delta relative to the shared base.
Prior work has demonstrated that such deltas can be compressed aggressively while still enabling accurate reconstruction of the fine-tuned model at inference time \citep{liu2024bitdelta}. However, they rely on coarse parametrizations that ignore variation in residual scales across rows or columns of weight matrices, leading to reconstruction errors that could be avoided with more structured representations. At the same time, introducing too much precision or auxiliary metadata risks negating the efficiency benefits.

This paper introduces a \emph{1-bit delta} representation—storing only the binary sign mask of the weight difference $\mathbf{B}=\mathrm{sign}(\mathbf{W_f}-\mathbf{W_b})$ and learning lightweight per-row/column scales—designed to balance those trade-offs: maintaining the simplicity and low storage overhead
of delta compression, while adding lightweight per-axis scaling to better capture the axis-specific
patterns in model weights. We show that this approach improves approximation quality at negligible
extra cost, enabling faster and more memory-efficient serving of many fine-tuned variants from a
single shared base model.

\section{Method}
We propose a parameter-efficient method for storing a fine-tuned model by leveraging its shared architecture with a base model. The core idea is to represent the output of fine-tuned weights as a sum of the base weights and a compressed residual term.

\begin{figure}[hbt!]
  \centering
  \includegraphics[width=\linewidth]{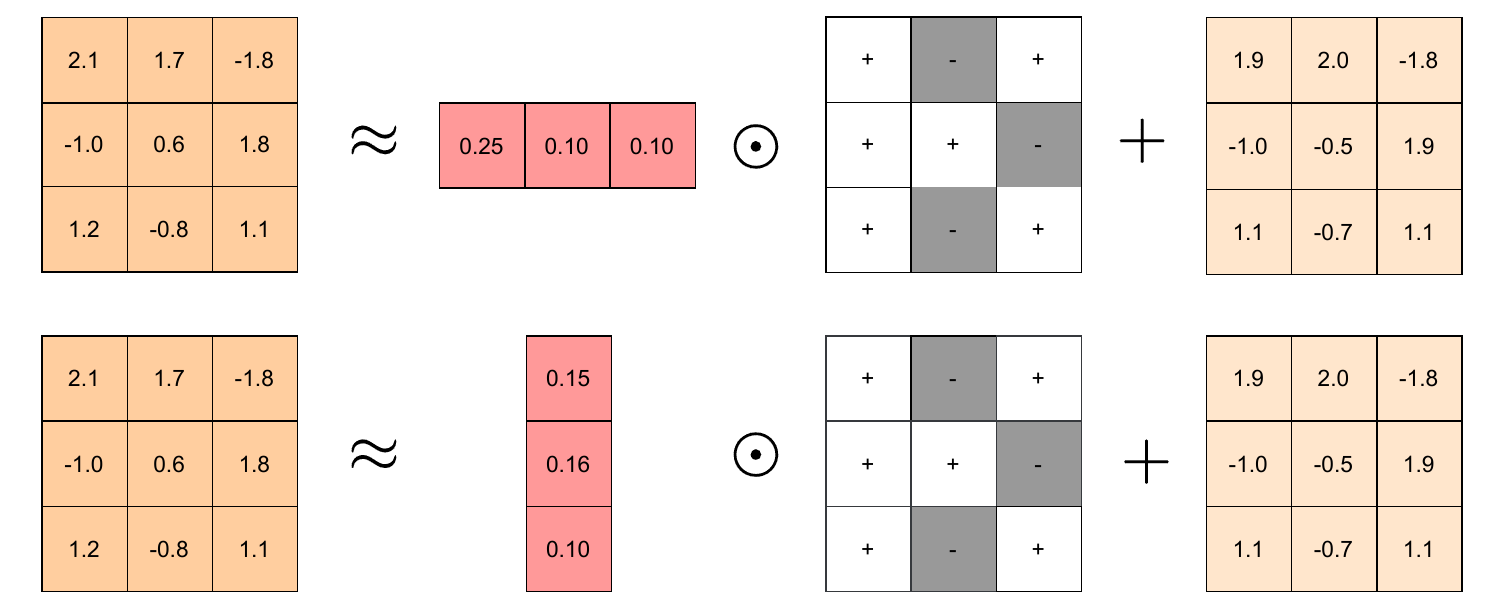}

  \caption{Approximating the fine-tuned weights $\mathbf{W_f}$ by $\mathbf{v} \odot \mathbf{B} + \mathbf{W_b}$: 
a compact 1-bit sign residual, where $\mathbf{v}$ is a vector, 
$\mathbf{B} \in \{-1,+1\}$ is the binary sign matrix, and $\mathbf{W_b}$ is the base weight matrix.}

  \label{fig:overview}
\end{figure}

Let a model be composed of $L$ layers. For layer $i$ we have base and fine-tuned weights $W_b^{(i)},W_f^{(i)}\!\in\!\mathbb{R}^{d_{\text{out}}\times d_{\text{in}}}$. We define
$\Delta \mathbf{W}^{(i)}\!=\!\mathbf{W_f}^{(i)}-\mathbf{W_b}^{(i)}$ the 1-bit sign mask
$\mathbf{B}^{(i)}=\mathrm{sign}(\Delta \mathbf{W}^{(i)})\in\{-1,+1\}^{d_{\text{out}}\times d_{\text{in}}}$.
After that we patch via a per-axis broadcasted scale
\[
\mathbf{\widehat{W}}^i = \mathbf{v}^{(i)} \odot \mathbf{B}^{(i)} + \mathbf{W_b}^{(i)}, \quad
\mathbf{v}^{(i)} \in 
\begin{cases}
\mathbb{R}^{1\times d_{\text{out}}} & \text{(row)},\\
\mathbb{R}^{d_{\text{in}}\times1}  & \text{(col)},
\end{cases}
\]

where $\odot$ replicates $\mathbf{v}^{(i)}$ by columns (row mode) or rows (col mode). The row or column mode is selected dynamically to best approximate the model’s fine-tuned layer output; see Fig.~\ref{fig:overview}.

This approach achieves significant compression. The storage cost per layer is reduced from floating-point weights to a single bitmask and a single vector. This enables the efficient storage of multiple fine-tuned models specialized for different tasks, all of which share the same underlying base weights. 

\paragraph{Prior evidence against weight reconstruction.}

The objective is not to recover the exact parameter values, but to preserve the function the network computes - i.e., to match outputs under realistic inputs. A line of works shows that minimizing weight-space error (e.g., round-to-nearest) is a weak surrogate for preserving model behavior:
(i) \citet{pmlr-v119-nagel20a}  demonstrate that round-to-nearest is suboptimal and introduce \emph{loss-aware} adaptive rounding that consistently outperforms weight-nearest at low bit widths;
(ii) \citet{frantar2023gptqaccurateposttrainingquantization} explicitly minimize \emph{layer-output} error (Hessian-aware) and report large gains over RTN on LLMs at 3--4 bits;
(iii) \citet{li2021brecqpushinglimitposttraining} formulate \emph{block reconstruction} of activations with a second-order analysis, enabling PTQ at 2 bits;
(iv) \citet{lin2024awqactivationawareweightquantization} argue that salient channels should be selected via \emph{activation} statistics rather than weights;
(v) \citet{pmlr-v202-xiao23c} argue that while weights are relatively straightforward to quantize compared to activations, the difficulty can be mitigated by rescaling weights to absorb part of the activation complexity.

\paragraph{Calibration cache, training, and stacking.}
For each target layer $i$, the vector $ \mathbf{v}^{(i)}$ is trainable while 
$\mathbf{W_b}^{(i)}$ and $\mathbf{B}^{(i)}$ are frozen at inference. We extract a small calibration set of 50 C4 \citep{raffel2020t5} samples and build a per-layer cache of $(\mathbf{X},\mathbf{Y})$ pairs:
$\mathbf{X}$ is the input that has to be passed to the $i$ layer of the compressed model (i.e., the output of the already-compressed stack up to layer $i\!-\!1$, immediately before entering layer $i$), and $\mathbf{Y}$ are the fine-tuned outputs of the original non-compressed finetuned  layer, while $\hat{\mathbf{Y}}$ denotes the output produced by compressed layer. We attach forward hooks to the teacher to collect $\mathbf{Y}$ and to the student to collect $\mathbf{X}$, store both as BF16 tensors. For each target layer $i$ we instantiate both axis variants and fit
only their scale vectors on the cache with an MSE objective, \[
\mathcal{L}_{\text{layer}}=\tfrac{1}{n}\big\|\,\mathbf{Y}-\mathbf{\hat{Y}}\,\big\|_2^2,
\]
using AdamW for 5 epochs under the same budget across
variants. The axis is selected by validation MSE on the held-out shard, and the original layer is replaced with the
better variant.  We sweep all
linear projections in attention and MLP blocks and install the selected  module
per layer, yielding a compressed student stacked on top of the shared base. Finally, we use an additional set of 150 C4 examples to jointly train all selected vectors on end-to-end objective, ensuring that the final stacked model accurately reproduces the teacher’s output.

\paragraph{Implementation remarks.}
We run on Llama-3.1-8B, using Llama-3.1-8B-Instruct as the teacher and Llama-3.1-8B as the student. Due to limited VRAM, we used two RTX 4090 GPUs and split fine-tuned compressed weights across devices. We cache teacher layer outputs (fine-tuned, \texttt{cuda:0}) and student inputs
(compressed, \texttt{cuda:1}) via forward hooks as detached BF16 clones stored on \texttt{cuda:1}.
Masks $\mathbf{B}^{(i)}$ stay packed end-to-end (1 bit along input axis), vectors $ \mathbf{v}^{(i)}$ are FP16, and
base weights are kept as $(\mathrm{in},\mathrm{out})$ BF16. We use non-blocking transfers and a single
\texttt{.to(device)} per module. The full algorithm can be seen in Algorithm \ref{alg:global-pipeline}.

\section{Experiments}

\label{sec:experiments}
\subsection{Setup}
We adopt a simple evaluation setting: Llama-3.1-8B as the base model and Llama-3.1-8B-Instruct as the fine-tuned target, evaluated zero-shot on ARC-Challenge, ARC-Easy \citep{clark2018arc}, HellaSwag \citep{zellers2019hellaswag}, PIQA \citep{bisk2020piqa}, and Winogrande \citep{sakaguchi2020winogrande}. All methods use the same calibration budget of 150 samples for end-to-end objective training and 50 samples used to train a vector that replicates the output of the real layer, drawn from C4 \citep{raffel2020t5}.

For our vector scales we use AdamW, learning rate \({1\times 10^{-4}}\), for five epochs, as the longer parameterization requires more steps to converge stably; BitDelta (scalar) uses the same pipeline but with a single scalar per matrix and one epoch for training.
We report zero-shot accuracy (\%) on the public test splits using the same prompt formatting across methods.

For additional descriptive analysis of the selected delta-quantization axis, see Appendix~\ref{app:delta-axis}; per-sub-type counts and layer-wise trends are shown in Figure~\ref{fig:axis_counts}.

\paragraph{Models and baselines.}
Baseline denotes the fine-tuned model without any delta compression.
BitDelta (scalar) is the 1-bit sign mask with a single learned scalar per matrix.
Our method is with a 1-bit sign mask and a learned per-row or per-column vector of scales. 

\subsection{Main results}
Table~\ref{tab:main-results} summarizes the \emph{zero-shot accuracy} on ARC-Challenge, ARC-Easy, HellaSwag, PIQA, and Winogrande, using several model pairs. Each pair consists of a base model and its fine-tuned target: Llama-3.1-8B / Llama-3.1-8B-Instruct \citep{grattafiori2024llama3herdmodels}, Qwen3-14B-Base / Qwen3-14B \citep{yang2025qwen3technicalreport}, and Phi-4 / Phi-4-Reasoning \citep{abdin2024phi4technicalreport}.

Vector (row/col) improves the average score over the baseline by \(0.97\) points and over BitDelta (scalar) by \(0.28\) points for the Llama pair, and by \(0.37\) points for the Qwen3 pair, where BitDelta performs \(0.29\) points below the baseline.

Gains are consistent on ARC-Challenge/Easy and Winogrande; HellaSwag is on par, while PIQA shows a small drop versus BitDelta. See Appendix~\ref{app:delta-axis} for a breakdown by module sub-type (Figure~\ref{fig:axis_counts}).

\paragraph{Storage and load-time.}
Our delta representation stores the fine-tuned model as a compact $\sim\!2.97$\,GB artifact on disk for the 8B setting (Table~\ref{tab:storage})—about $5.24\times$ smaller than a full FP16 checkpoint. Under identical allocator/seeds and cold-start conditions on Llama-3.1-8B, the average load time over 10 runs to apply the vector–delta on top of the base is $0.80$\,s, whereas loading the entire fine-tuned FP16 checkpoint takes $2.08$\,s. Thus the delta path uses less per-model load time for a much smaller on-disk and transfer footprint per-model. This is especially useful when maintaining or hot-swapping many fine-tuned versions of a given base model.

\begin{table}[t]
\centering
\small
\caption{Zero-shot accuracy (\%) after calibrating on 150 samples from C4. Vector scales are trained for five epochs with learning rate \(1\mathrm{e}{-5}\); Single scalar per matrix uses the same setup with one epoch.}
\begin{tabular}{llcccccc}
\toprule
Model & Method & ARC-C & ARC-E & HellaSwag & PIQA & Winogrande & Avg \\
\midrule
\multirow{3}{*}{\shortstack[l]{Llama-3.1-8B-Instruct\\ \scriptsize(Base: Llama-3.1-8B)}}
& Baseline           & 51.70 & 81.81 & 59.06 & 79.86 & 73.87 & 69.26 \\ &
  BitDelta (scalar)  & 52.55 & 82.32 & 59.73 & \textbf{81.22} & 73.95 & 69.95 \\ &
 Vector (row/col)   & \textbf{53.58} & \textbf{82.99} & \textbf{59.78} & 80.63 & \textbf{74.19} & \textbf{70.23} \\

\midrule
\multirow{3}{*}{\shortstack[l]{Qwen3-14B\\ \scriptsize(Base: Qwen3-14B-Base)}}
 & Baseline           & 58.87 & 84.09 & 60.89 & 80.09 &  72.77 & 71.34 \\ &
  BitDelta (scalar)  & \textbf{58.70} & 84.13 & 59.88 & 79.38 & \textbf{73.16} & 71.05 \\ &
 Vector (row/col)   & \textbf{58.70} & \textbf{84.34} & \textbf{62.07} & \textbf{80.52} & 72.93 & \textbf{71.71} \\

 \midrule
\multirow{3}{*}{\shortstack[l]{Phi-4-reasoning\\ \scriptsize(Base: Phi-4)}}
 & Baseline           & 55.72 & 83.29 & 59.01 & 80.63 &  75.61 & 70.74 \\ &
  BitDelta (scalar)  & 55.46 & \textbf{83.54} & \textbf{59.49} & 80.74 & 76.09 & \textbf{71.06} \\ &
 Vector (row/col)   & \textbf{55.63} & 82.95 & 59.35 & \textbf{80.85} & \textbf{76.24} & 71.00 \\
\bottomrule
\end{tabular}

\label{tab:main-results}
\end{table}

\begin{table}[t]
\centering
\caption{Checkpoint sizes 
}

\small
\begin{tabular}{llccc}
\toprule
Model & Artifact & Size (MB)   & vs.\ FP16 weights  \\
\midrule
\multirow{2}{*}{\shortstack[l]{Llama-3.1-8B-Instruct\\ \scriptsize(Base: Llama-3.1-8B)}} &
BitDelta (scalar) & 2974   & \(\approx 5.25\times\) smaller  \\ &
Vector (row/col) &  2980   & \(\approx 5.24\times\) smaller \\

\midrule
\multirow{2}{*}{\shortstack[l]{Qwen3-14B\\ \scriptsize(Base: Qwen3-14B-Base)}} &
BitDelta (scalar) & 4775   & \(\approx 6.18\times\) smaller  \\ &
Vector (row/col) &  4774   & \(\approx 6.19\times\) smaller \\

\midrule
\multirow{2}{*}{\shortstack[l]{Phi-4-reasoning\\ \scriptsize(Base: Phi-4)}} &
BitDelta (scalar) & 3760   & \(\approx 7.80\times\) smaller  \\ &
Vector (row/col) &  3768  & \(\approx 7.78\times\) smaller \\

\bottomrule
\end{tabular}

\label{tab:storage}
\end{table}

\section{Limitations}
\label{sec:limitations}

Our gains rely on the anisotropy of the task-induced deltas $\mathbf{\Delta W}$ across rows/columns.
If a layer’s delta is nearly isotropic, a single global scale can match quality while avoiding the metadata and compute introduced by per-row/column vectors.

We fix $\mathbf{B}\!\in\!\{-1,+1\}^{d_{\text{in}}\times d_{\text{out}}}$ at 1\,bit per entry.
This forbids explicit zeros/sparsity and can propagate noise for very small-magnitude entries unless one adds debiasing or confidence filtering.
Consequently, the patch is dense and incurs slight additional MACs (extra steps) and memory overhead compared to a pure binary (sign-only) matrix.

Vector scales are learned with an activation-aware objective using a small calibration set to estimate $C_x$.
Distribution shift between calibration and deployment may reduce effectiveness; larger calibration improves robustness but increases preparation time and memory. 

We patch linear projections (attention and MLP).
We do not modify normalizations, biases or embeddings; if task-specific changes concentrate there, our method may yield limited benefits. The sign mask $\mathbf{B}$ is fixed and we do not learn signs or structure.
At aggressive bit budgets, learning $\mathbf{B}$ may be beneficial for downstream performance.

We add all residual terms at once for a selected fine-tuned model, yielding inference identical to FP16 weights with no further runtime overhead. An alternative on-the-fly variant could apply them dynamically in each forward pass and avoid switch costs, but would introduce runtime overhead unless supported by fused GEMM kernels.

\section{Conclusion}
We introduced a 1-bit delta scheme with lightweight per-axis (row/column) FP16 scales learned via output matching. Empirically, across five zero-shot benchmarks, the method consistently improves over both the uncompressed baseline and a scalar 1-bit delta variant. Averaged over all tasks, vector achives 70.23, 71.71, and 71.00, compared to 69.95, 71.05, and 71.06 for scalar method, and 69.26, 71.34, and 70.74 for the uncompressed models. Limitations include layers with near-isotropic deltas and reliance on small calibration sets. Future work includes blockwise per-group scaling, learning the sign structure, INT4/FP8 co-design, and broader multi-tenant evaluations.

Our method delivers higher average accuracy than both the baseline and the scalar BitDelta while preserving the same storage efficiency.
From a systems perspective, our loader reduces cold-start latency. Overall, vector scales provide a better match to the anisotropy of task deltas at negligible extra storage cost.

\section{Acknowledgements}
This research was partially supported by the Bulgarian
National Program ”Education with Science”

\bibliographystyle{plainnat}
\bibliography{refs}
\FloatBarrier
\newpage

\appendix
\section{Additional analysis of delta-quantization axis}
\label{app:delta-axis} 
This appendix provides descriptive statistics for the learned choice of the delta-quantization axis (\texttt{row} vs.\ \texttt{column}) across module sub-types and depth.

\begin{figure}[hbt!]
  \centering
  \includegraphics[width=0.9\linewidth]{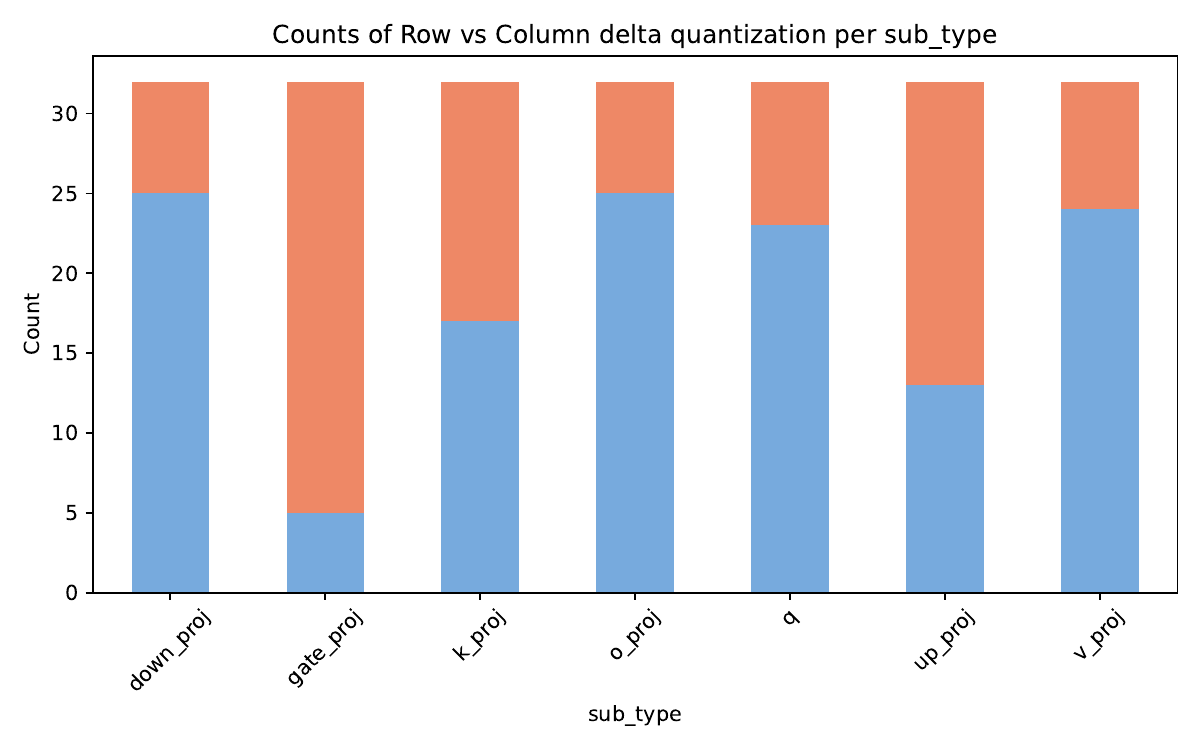}
  \caption{counts of \texttt{row} vs.\ \texttt{column} delta-quantization per sub\_type (row in blue, column in red).}
  \label{fig:axis_counts}
\end{figure}

\paragraph{Counts by sub-type.}
Figure~\ref{fig:axis_counts} summarizes how often each sub-type selects a \texttt{row} or a \texttt{column} axis for delta quantization.
Overall, attention projections (\texttt{q}, \texttt{v\_proj}, \texttt{o\_proj}) and the MLP \texttt{down\_proj} tend to prefer \texttt{row}, while \texttt{gate\_proj} and \texttt{up\_proj} show a stronger \texttt{column} preference, with \texttt{k\_proj} being more mixed.
These tendencies are consistent with the differing input/output aspect ratios of the corresponding weight matrices.

\begin{algorithm}[ht]
\caption{Layer-wise compression pipeline with cached activations}
\label{alg:global-pipeline}
\begin{algorithmic}[1]
\Require Teacher model $W_f$ on \texttt{cuda:0}, student model $\widehat{W}$ on \texttt{cuda:1} (initialized from the base model), train loader $\mathcal{D}_{\mathrm{tr}}$, val loader $\mathcal{D}_{\mathrm{val}}$, set of target layer names $\mathcal{L}$ (e.g., all MLP/self\_attn proj layers), train steps $T$, eval steps $E$
\Ensure Student $\widehat{W}$ where each $L \in \mathcal{L}$ is replaced by the better of \textsc{Row}/\textsc{Col}

\State Initialize an empty map $\mathsf{Cache}$

\State \textbf{Stage 1: build and store calibration caches (Algorithm~\ref{alg:calib-hooks-v2})}
\For{each layer name $L \in \mathcal{L}$}
  \State $(X_{\mathrm{tr}}^{(L)}, Y_{\mathrm{tr}}^{(L)}), (X_{\mathrm{val}}^{(L)}, Y_{\mathrm{val}}^{(L)}) \gets$
    Alg.~\ref{alg:calib-hooks-v2} with $(W_f, \widehat{W}, L, \mathcal{D}_{\mathrm{tr}}, \mathcal{D}_{\mathrm{val}}, T, E)$
  \State $\mathsf{Cache}[L] \gets \big(X_{\mathrm{tr}}^{(L)}, Y_{\mathrm{tr}}^{(L)}, X_{\mathrm{val}}^{(L)}, Y_{\mathrm{val}}^{(L)}\big)$
\EndFor

\State

\State \textbf{Stage 2: per-layer compression using Algorithm~\ref{alg:compress-layer-v2}}
\State \textbf{Stage 3: train all vectors using Algorithm~\ref{alg:e2e-train-all}}
\State \textbf{Stage 4: save vectors using Algorithm~\ref{alg:apply-saved}}
\end{algorithmic}
\end{algorithm}

\begin{algorithm}[t]
\caption{End-to-end calibration of all row/col vectors before deployment}
\label{alg:e2e-train-all}
\begin{algorithmic}[1]
\Require Teacher model $W_f$, compressed student model $\widehat{W}$  with row/col modules and scaling vectors $v$, train loader $\mathcal{D}_{\mathrm{tr}}$, val loader $\mathcal{D}_{\mathrm{val}}$, epochs $K$, learning rate $\eta$

\State Collect all learnable scaling vectors 
  $\theta_v \gets v $
\State Initialize AdamW optimizer on $\theta_v$ with learning rate $\eta$

\State

\State \textbf{Train (end-to-end matching of logits)}
\For{$k = 1$ to $K$}
  \For{each batch $b$ in $\mathcal{D}_{\mathrm{tr}}$}
    \State Move $b$ to \texttt{cuda:1}
    \State \textbf{Student logits:} $\ell_{\mathrm{pred}}\ \gets \widehat{W}(b)$
    \State \textbf{Teacher logits:} $\ell \gets {W}(b)$
  
    \State $L_{\mathrm{e2e}} \gets \|\ell - \ell_{\mathrm{pred}}\|_2^{2}$
    \State Backpropagate $L_{\mathrm{e2e}}$ only through $\theta_v$; optimizer step
  \EndFor
\EndFor

\State

\State \textbf{Optional: evaluate final end-to-end loss}
\State $L_{\mathrm{end}} \gets$ Alg.~\ref{alg:end-loss} on $\widehat{W}$ and $\mathcal{D}_{\mathrm{val}}$ 
\State
\end{algorithmic}
\end{algorithm}

\begin{algorithm}[t]
\caption{Register and use forward hooks to build calibration caches for a layer $L$}
\label{alg:calib-hooks-v2}
\begin{algorithmic}[1]
\Require Teacher model $W_f$ on \texttt{cuda:0}, student $\widehat{W}$ on \texttt{cuda:1}, target layer name $L$, train loader $\mathcal{D}_{\mathrm{tr}}$, val loader $\mathcal{D}_{\mathrm{val}}$, train steps $T$, eval steps $E$
\Ensure Caches $(X_{\mathrm{tr}},Y_{\mathrm{tr}})$ and $(X_{\mathrm{val}},Y_{\mathrm{val}})$ on \texttt{cuda:1}
\State Initialize empty maps $\textsc{Inputs}[L]$, $\textsc{Outputs}[L]$  \Comment{device=\texttt{cuda:1}, dtype=BF16}
\State $h_{\text{out}} \gets$ register forward hook on $W_f[L]$ that appends \emph{detached BF16 output} to $\textsc{Outputs}[L]$ on \texttt{cuda:1}
\State $h_{\text{in}} \gets$ register forward hook on $\widehat{W}[L]$ that appends \emph{detached BF16 input} to $\textsc{Inputs}[L]$ on \texttt{cuda:1}
\For{$t=1$ to $T$} \Comment{build train cache}
  \State Fetch batch $b \leftarrow \mathcal{D}_{\mathrm{tr}}$
  \State Run $W_f$ on $b$ (moved to \texttt{cuda:0}, no grad) \Comment{fills $\textsc{Outputs}[L]$}
  \State Run $\widehat{W}$ on $b$ (moved to \texttt{cuda:1}, no grad) \Comment{fills $\textsc{Inputs}[L]$}
\EndFor
\For{$e=1$ to $E$} \Comment{build val cache}
  \State Repeat the two forwards with $\mathcal{D}_{\mathrm{val}}$
\EndFor
\State Remove hooks $h_{\text{out}}$, $h_{\text{in}}$
\State $X_{\mathrm{tr}},Y_{\mathrm{tr}} \gets$ first $T$ items of $\textsc{Inputs}[L],\textsc{Outputs}[L]$
\State $X_{\mathrm{val}},Y_{\mathrm{val}} \gets$ last $E$ items of $\textsc{Inputs}[L],\textsc{Outputs}[L]$
\State \Return $(X_{\mathrm{tr}},Y_{\mathrm{tr}}),(X_{\mathrm{val}},Y_{\mathrm{val}})$
\end{algorithmic}
\end{algorithm}

\begin{algorithm}[t]
\caption{Train per-row/column scaling vectors via activation matching}
\label{alg:train-delta-v2}
\begin{algorithmic}[1]
\Require Compressed layer $M$ (\textsc{row} or \textsc{col}) with learnable \textit{v}, train cache $(X_{\mathrm{tr}},Y_{\mathrm{tr}})$, val cache $(X_{\mathrm{val}},Y_{\mathrm{val}})$, epochs $K$, learning rate $\eta$

\State Initialize AdamW on \textit{v} only with LR $\eta$; optional cosine scheduler over $K$ epochs
\For{$k=1$ to $K$}
  \State \textbf{Train:} For each minibatch $(x,y)\in (X_{\mathrm{tr}},Y_{\mathrm{tr}})$:
    \State \hspace{1em} $y_{\mathrm{pred}} \gets M(x)$ 
    \State \hspace{1em} $L \gets \big\|\,y-y_{\mathrm{pred}}\,\big\|_2^2$; backprop only through \textit{v}; optimizer step; scheduler step
\EndFor
\State \textbf{Validate:} $L_{\mathrm{val}} \gets$ mean of $\|Y_{\mathrm{val}} - $M$(X_{\mathrm{val}})\|_2^2$ over $(X_{\mathrm{val}},Y_{\mathrm{val}})$ (no grad)
\State \Return $(\textit{v}, L_{\mathrm{val}})$
\end{algorithmic}
\end{algorithm}

\begin{algorithm}[t]
\caption{End-to-end validation loss (student vs.\ cached teacher logits)}
\label{alg:end-loss}
\begin{algorithmic}[1]
\Require Student model $\widehat{W}$ on \texttt{cuda:1}, val loader $\mathcal{D}_{\mathrm{val}}$, cached teacher logits $\{\ell^{\ast}_t\}$ aligned by batch
\Ensure Validation loss $L_{\mathrm{end}}$
\State $L_{\mathrm{end}}\gets 0$, $n\gets 0$
\For{first $N$ batches $b$ in $\mathcal{D}_{\mathrm{val}}$}
  \State Move $b$ to \texttt{cuda:1}, run $\widehat{W}$ under AMP to get logits $\ell$
  \State $L_{\mathrm{end}} \gets L_{\mathrm{end}} + \|\ell - \ell^{\ast}_b\|_2^2$; $n \gets n+1$
\EndFor
\State \Return $L_{\mathrm{end}}/n$
\end{algorithmic}
\end{algorithm}

\begin{algorithm}[t]
\caption{Per-layer compression with Row/Col selection by end loss}
\label{alg:compress-layer-v2}
\begin{algorithmic}[1]
\Require Base weight $W_b^{(L)}$, fine-tuned $W_f^{(L)}$, layer name $L$, loaders $\mathcal{D}_{\mathrm{tr}},\mathcal{D}_{\mathrm{val}}$
\Ensure Replace layer $L$ with the better of \textsc{Row}/\textsc{Col}
\State $\Delta W \gets W_f^{(L)}-W_b^{(L)}$;\quad $B \gets \textsc{Pack}(\mathrm{sign}(\Delta W)^\top)$
\State $(X_{\mathrm{tr}},Y_{\mathrm{tr}}),(X_{\mathrm{val}},Y_{\mathrm{val}}) \gets$ Alg.~\ref{alg:calib-hooks-v2} for $L$
\State Build \textbf{Col} module $M_{\text{col}}(B,v)$ with $v_c \gets \mathrm{mean}(|\Delta W|,\text{axis}=1)$; train via Alg.~\ref{alg:train-delta-v2} with LR $1\!\times\!10^{-4}$, epochs $5$
\State $E_{\text{col}} \gets$ Alg.~\ref{alg:end-loss} on $\widehat{W}$ after swapping in $M_{\text{col}}$
\State Build \textbf{Row} module $M_{\text{row}}(B,v_r)$ with $v_r \gets \mathrm{mean}(|\Delta W|,\text{axis}=0)$; train via Alg.~\ref{alg:train-delta-v2} with LR $1\!\times\!10^{-4}$, epochs $5$
\State $E_{\text{row}} \gets$ Alg.~\ref{alg:end-loss} on $\widehat{W}$ after swapping in $M_{\text{row}}$
\If{$E_{\text{row}} \le E_{\text{col}}$} \State \textsc{ReplaceLayer}$(L \leftarrow M_{\text{row}})$
\Else \State \textsc{ReplaceLayer}$(L \leftarrow M_{\text{col}})$
\EndIf
\end{algorithmic}
\end{algorithm}

\begin{algorithm}[t]
\caption{Model-wide application from a saved delta file (row/col-aware)}
\label{alg:apply-saved}
\begin{algorithmic}[1]
\Require Student model $\widehat{W}$, delta dict $\mathsf{diff}$ (keys: \texttt{.mask\_row}, \texttt{.coeff\_row}, \texttt{.mask\_col}, \texttt{.coeff\_col})
\ForAll{modules $(\text{name},\text{mod})$ in $\widehat{W}$ where $\text{NameContains}(\text{name},\{\text{mlp},\text{self\_attn}\})$ and $\text{NameContains}(\text{subname},\{\text{proj}\})$}
  \If{$\mathsf{diff}$ has $\text{name}{+}$.mask\_row} \State \textsc{CompressLayerRow}$(\text{name}, \mathsf{diff})$ 
  \ElsIf{$\mathsf{diff}$ has $\text{name}{+}$.mask\_col} \State \textsc{CompressLayerCol}$(\text{name}, \mathsf{diff})$
  \EndIf
\EndFor

\end{algorithmic}
\end{algorithm}

\end{document}